\documentclass[10pt,journal,compsoc]{IEEEtran}
\usepackage{booktabs}
\usepackage{graphicx}
\usepackage{graphics}
\usepackage{placeins}
\usepackage{float}
\usepackage{subfigure}
\usepackage{amsmath}
\usepackage{url}

\usepackage[table]{xcolor}
\usepackage{collcell}
\usepackage{hhline}
\usepackage{pgf}
\usepackage{multirow}
\usepackage{tikz}
\usetikzlibrary{calc}
\usepackage {color, graphicx}
\usepackage {subfigure}
\usetikzlibrary{decorations.pathmorphing}
\usetikzlibrary {patterns,shapes,shadows,decorations.pathreplacing}
\usepackage{amssymb}

\ifCLASSOPTIONcompsoc
  \usepackage[nocompress]{cite}
\else
  \usepackage{cite}
\fi
\ifCLASSINFOpdf
\else
\fi

\newcommand{\etal}{\mbox{\emph{et al.}}}
\newcommand{\revision}[1]{#1}

\newcommand\copyrighttext{%
  \footnotesize \bf © 2019 IEEE.  Personal use of this material is permitted.  Permission from IEEE must be obtained for all other uses, in any current or future media, including reprinting/republishing this material for advertising or promotional purposes, creating new collective works, for resale or redistribution to servers or lists, or reuse of any copyrighted component of this work in other works.}
\newcommand\copyrightnotice{%
\begin{tikzpicture}[remember picture,overlay]
\node[anchor=south,yshift=0pt] at (current page.south) {\fbox{\parbox{\dimexpr\textwidth-\fboxsep-\fboxrule\relax}{\copyrighttext}}};
\end{tikzpicture}%
}

\begin{document}

%
\title{Hydra: an Ensemble of Convolutional Neural Networks for Geospatial Land Classification}
%
%
%
%

\author{ Rodrigo~Minetto,
         Maur\'icio~Pamplona~Segundo,~\IEEEmembership{Member,~IEEE,}
         Sudeep~Sarkar,~\IEEEmembership{Fellow,~IEEE}
\IEEEcompsocitemizethanks{
\IEEEcompsocthanksitem R. Minetto is with Universidade Tecnol\'{o}gica Federal do Paran\'{a} (UTFPR), Brazil. E-mail: rodrigo.minetto@gmail.com
\IEEEcompsocthanksitem M. P. Segundo is with Universidade Federal 
da Bahia (UFBA), Brazil. E-mail: mauriciops@ufba.br
\IEEEcompsocthanksitem S. Sarkar is with Department of Computer Science and Engineering, University of South Florida (USF), Tampa, FL, USA. E-mail: sarkar@usf.edu
\IEEEcompsocthanksitem The research in this paper was conducted while the authors were at the Computer Vision and Pattern Recognition Group, USF.}
}
%
%

\markboth{To appear in IEEE Transactions on Geoscience and Remote Sensing,~2019}%
{Minetto \MakeLowercase{\textit{et al.}}: Hydra: an Ensemble of Convolutional Neural Networks for Geospatial Land Classification}
\maketitle


%

\copyrightnotice

\begin{abstract}
We describe in this paper Hydra, an ensemble of convolutional neural networks (CNN) for geospatial land classification. The idea behind Hydra is to create an initial CNN that is coarsely optimized but provides a good starting pointing for further optimization, which will serve as the Hydra's body. Then, the obtained weights are fine-tuned multiple times with different augmentation techniques, crop styles, and classes weights to form an ensemble of CNNs that represent the Hydra's heads. By doing so, we prompt convergence to different endpoints, which is a desirable aspect for ensembles. With this framework, we were able to reduce the training time while maintaining the classification performance of the ensemble. We created ensembles for our experiments using two state-of-the-art CNN architectures, ResNet and DenseNet. We have demonstrated the application of our Hydra framework in two datasets, FMOW and NWPU-RESISC45, achieving results comparable to the state-of-the-art for the former and the best reported performance so far for the latter. Code and CNN models are available at \url{https://github.com/maups/hydra-fmow}.
\end{abstract}


\section{Introduction}
\label{sec:introduction}

\begin{figure}[!htb]
  \begin{tikzpicture}
   \tikzset{blockA/.style={draw, ellipse, text centered, drop shadow, fill=white, text width=2.3cm, minimum height=0.4cm, inner sep=0pt}}
   \tikzset{blockB/.style={draw, ellipse, text centered, drop shadow, fill=white, text width=1.9cm, minimum height=0.4cm, inner sep=0pt}}
     
  \path[->](1.2,12.2) node[]  {\small Spectral: 4/8 band multispectral images};   
     
  \draw(1.2,9.9) node[inner sep=0pt] (img1) {
     \includegraphics[width=5.0cm]{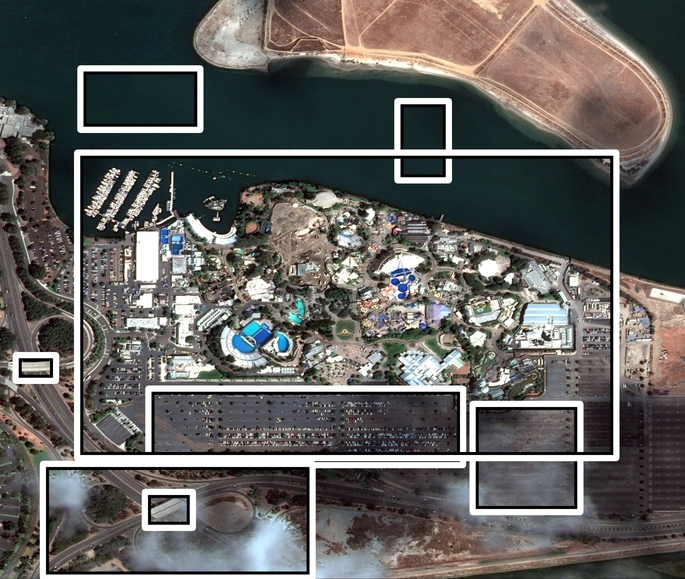}   
  };
    
  \node[inner sep=2pt, rotate=45] (dots) at (+0.1,7.55) {\Large \ldots};
  
   \draw(-1.2,5.1) node[inner sep=0pt, opacity=0.2] (img_tp2) {
      \includegraphics[width=5.0cm]{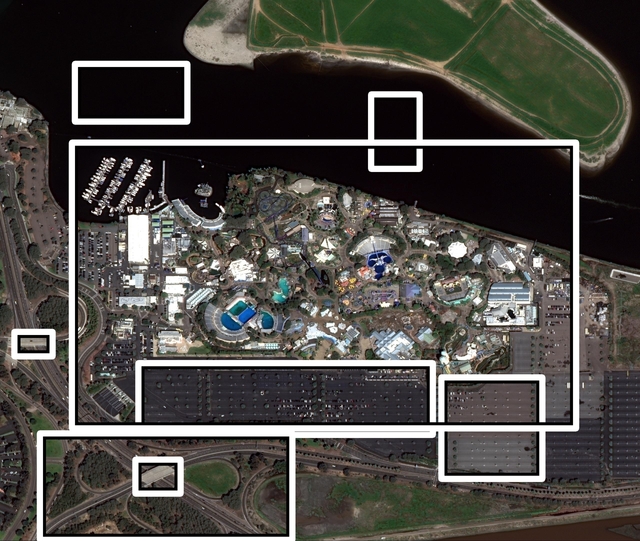}   
   };   
     
   \draw(-1.6,4.8) node[inner sep=0pt, opacity=0.2] (img_tp1) {
      \includegraphics[width=5.0cm]{figures/image_0004691_5_rgb.jpg}   
   };  
     
   \draw(-2.0,4.4) node[inner sep=0pt] (img2) {
      \includegraphics[width=5.0cm]{figures/image_0004691_5_rgb.jpg}   
   };

   \draw [->] (-4.7,6.7) -- (-3.0,8.2) node [midway,fill=white,inner sep=1pt] {\small Temporal};    
   \draw [->] (-4.7,6.7) -- (0.6,6.7) node [midway,fill=lightgray=0.9,inner sep=1pt] {\small Spatial}; 
   \draw [->] (-4.7,6.7) -- (-4.7,2.2) node [midway, rotate=90, fill=white,inner sep=1pt] {\small Spatial};

   \node[draw,align=left,rounded corners=5pt,drop shadow,fill=white] (textc) at (-2.7,10.2) {\baselineskip=10pt \footnotesize Textual metadata \\ \tiny \noindent\rule{1cm}{0.4pt} \\\footnotesize gsd: 0.44421135..\\ \footnotesize cloud cover: 25\\ \footnotesize date: 2013-08-25 \\ \footnotesize time 18:47:39 \\ \footnotesize utm: 11S\\ \footnotesize country: USA\\ ...};
   
   \node[draw,align=left,rounded corners=5pt,drop shadow,fill=white] (textc) at (+1.9,4.7) {\baselineskip=10pt \footnotesize Textual metadata \\ \tiny \noindent\rule{1cm}{0.4pt} \\\footnotesize gsd: 0.57625807..\\ \footnotesize cloud cover: 0\\ \footnotesize date: 2017-03-02 \\ \footnotesize time 18:21:45 \\ \footnotesize utm: 11S\\ \footnotesize country: USA\\ ...};
   
  \end{tikzpicture} 
  \caption{Typical data associated with aerial images for land use classification. It provides spatial, temporal, spectral and metadata information together with the annotation of target locations (example taken from the FMOW dataset~\cite{fmow2017}).} 
  \label{fig:fmow_spatial_temp}
\end{figure}

Land use is a critical piece of information for a wide range of applications, from humanitarian to military purposes. For this reason, automatic land use classification from satellite images has been drawing increasing attention from academia, industry and government agencies~\cite{fmow2017}. This research problem consists in classifying a target location in a satellite image as one of the classes of interest or as none of them. It is also common to have metadata associated with these images. Figure~\ref{fig:fmow_spatial_temp} illustrates a typical input data used for land use classification.

There are many factors that make the land classification problem very challenging:
\begin{itemize}
\item {\bf Clutter:} satellite images cover a large piece of land and may include a variety of elements ({\it e.g.}, objects, buildings, and vegetation), making it hard to classify a target place.
\item {\bf Viewpoint:} aerial images can be acquired from different angles depending on the satellite location, which could considerably change the appearance of the imaged content.
\item {\bf Occlusion:} some parts of a target place may not be visible in aerial images due to viewpoint variations, cloud cover or shadows.
\item {\bf Time:} variations over time affect the appearance in different ways. Short-term temporal variations include illumination changes and movable objects, such as vehicles and temporary facilities, while long-term variations consist of major topographic changes caused by construction, weather, among others.
\item {\bf Scale:} there is a huge variation in size between different target types ({\it e.g.}, airport versus swimming pool) and sometimes even between multiple instances of the same target type ({\it e.g.}, a parking lot), which is not easy to process with a single approach.
\item {\bf Functionality:} sometimes targets with similar appearance may have completely different functions. For instance, a similar building structure could be used as an office, a police station or an educational facility.
\end{itemize}


State-of-the-art methods tackle these difficulties by training a single Convolutional Neural Network (CNN) classifier over a large dataset of satellite images seeking to learn a generalizable model~\cite{fmow2017,Nogueira2017,7934005,7891544}, which were shown to outperform earlier works based on handcrafted features. Even though it is well known that ensembles of classifiers improve the performance of their individual counterparts~\cite{Dietterich2000}, this solution was not exploited by these works, probably due to its high training cost for large datasets. However, ensembles tackle one of the most common problems in multiclass classification, which is the existence of several critical points in the search space (saddle points and plateaus) that prioritize some classes over others and the eventual absence of a global minimum within the classifier search space. An ensemble ends up expanding this space by combining multiple classifiers that converged to different endpoints and reaches a better global approximation.

\begin{figure}[!ht]
\centering
\includegraphics[width=0.40\textwidth]{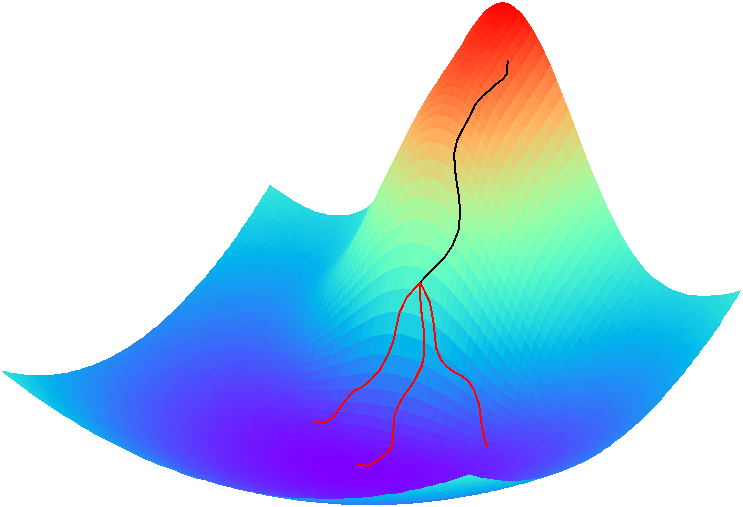}
\caption{Illustration of the optimization process in the Hydra framework. The black line represents the coarse parameter optimization that forms the Hydra's body. The red lines represent the Hydra's heads, which are obtained after fine tuning the body parameters multiple times seeking to reach different endpoints.}
\label{fig:hydra_opt}
\end{figure}

\revision{As our main contribution, we present a faster way of creating ensembles of CNNs for land use classification in satellite images, which we called Hydra for its similarity in shape to the namesake mythical creature (see Figure~\ref{fig:hydra_opt}). The idea behind Hydra is to create an initial CNN that is coarsely optimized but provides a good starting pointing for further optimization, which will serve as the Hydra's body. Then, the obtained weights are fine-tuned multiple times to form an ensemble of CNNs that represent the Hydra's heads. We take certain precautions to avoid affecting the diversity of the classifiers, which is a key factor in ensemble performance. By doing so, we were able to cut the training time approximately by half while maintaining the classification performance of the ensemble. Figure~\ref{fig:hydra_opt} illustrates this process, in which a black line represents the optimization of the body and the red lines represent the heads reaching different endpoints. To stimulate convergence to different endpoints during training and thus preserve diversity, we exploit different strategies, such as online data augmentation, variations in the size of the target region, and changing image spectra.}

\section{Related work}
\label{sec:related}

There is an extensive literature on algorithms for geospatial land image processing --- often called remote sensing image processing. This broad and active field of research has many branches, such as semantic segmentation~\cite{8068943,8393448}, target localization~\cite{7560644}, and region classification~\cite{7959553, 7936433, Nogueira2017, 7934005, 8252784}. 
In the domain of this work, there are different publicly available datasets for land use/land cover classification that can be used as performance benchmarks. Initial datasets used in the literature, such as UC Merced Land Use Dataset~\cite{Yang2010} (UCMerced, 21 classes, 2100 images), WHU-RS19~\cite{Dai2011} (19 classes, 950 images), Brazilian Coffe Scenes~\cite{Nogueira2017} (BCS, 2 class, 2876 images) and Aerial Image Dataset~\cite{7907303} (AID, 30 classes, 10000 images), had a few images and do not necessarily represent a challenging task ({\it i.e.}, reported results reach a nearly perfect score). Recently, two large datasets were created and yield a much harder classification problem: the Functional Map of the World~\cite{fmow2017} (FMOW, 63 classes, 470086 images) and the NWPU-RESISC45~\cite{7891544} (45 classes, 31500 images) datasets.


Given the results obtained so far for the aforementioned datasets, there is no doubt that deep CNNs are becoming more and more popular for land use classification. Nogueira~\emph{et al.}~\cite{Nogueira2017} compared different state-of-the-art CNN architectures, such as AlexNet~ \cite{Krizhevsky2017}, GoogleNet~\cite{43022}, and VGG~\cite{Simonyan2015}, and evaluated different training setups, such as using ImageNet weights as starting point and Support Vector Machines (SVM) for classification. They reported results for UCMerced, WHU-RS19 and BCS datasets, which were very successful when compared to many other approaches based on handcrafted features. Chaib~\emph{et al.}~\cite{7934005} improved regular VGG networks~\cite{Simonyan2015} by fusing the output of the last two fully connected layers. The work was validated using UCMerced, WHU-RS19, and AID. We also have CNN-based results reported by Christie~\etal~\cite{fmow2017} and Cheng~\etal~\cite{7891544} for the FMOW and NWPU-RESISC45 datasets, respectively, as later shown in Tables~\ref{table:fmow_performance}~and~\ref{table:nwpu-resisc45_performance}. Thus, CNN classifiers were definitely the best choice for our Hydra framework.

Other interesting works based on CNNs seek to extract a semantic segmentation from remote sensing images ({\it i.e.}, classification at the pixel level). This task is more challenging than region classification and suffers from the absence of large datasets. Xu~\emph{et al.}~\cite{8068943} combined hyperspectral images with terrain elevation acquired by a Light Detection And Ranging (LiDAR) sensor by using a two-branch CNN, one for each imaging modality. Each branch receives a small image patch around a pixel of its respective modality and extracts discriminative features from it. These features are then concatenated and classified to define the label of this specific pixel. Tao~\emph{et al.}~\cite{8013921} improved this idea by using Deconvolutional Neural Networks~\cite{6126474} --- an approximation of the inverse of a CNN --- to design an encoder-decoder architecture that receives a large satellite image patch as input and outputs a label for all its pixels at once. They also train part of their network in an unsupervised manner to ease the lack of labeled training data. Although we do not conduct any experiments on the semantic segmentation task, we believe it could also benefit from the Hydra framework proposed in this work.

\section{The Hydra Framework}
\label{sec:method}

Ensembles of learning algorithms have been effectively used to improve the classification performance in many computer vision problems~\cite{7299170,Simonyan2015,7917252}. The reasons for that, as pointed out by Dietterich~\cite{Dietterich2000} are: 1) the training phase does not provide sufficient data to build a single best classifier; 2) the optimization algorithm fails to converge to the global minimum, but an ensemble using distinct starting points could better approximate the optimal result; 3) the space being searched may not contain the optimal global minimum, but an ensemble may expand this space for a better approximation.

Formally, a supervised machine learning algorithm receives a list of training samples $\{(x_1, y_1), (x_2, y_2), \dots , (x_m, y_m)\}$ drawn from some unknown function $y = f(x)$ --- where $x$ is the feature vector and $y$ the class --- and searches for a function $h$ through a space of functions $\mathcal{H}$, also known as hypotheses, that best matches $y$. As described by Dietterich~\cite{Dietterich2000}, an ensemble of learning algorithms build a set of hypotheses $\{h_1, h_2, \dots, h_k\} \in \mathcal{H}$ with respective weights $\{w_1, w_2, \dots, w_k\}$, and then provides a classification decision $\bar{y}$ through hypotheses fusion. Different fusion approaches may be exploited, such as weighted sum $\bar{y} = \sum_{i = 1}^{k} w_i h_i (x)$, majority voting, etc.

Hydra is a framework to create ensembles of CNNs. As pointed out by Sharkey~\cite{Sharkey1999}, a neural network ensemble can be built by varying the initial weights, varying the network architecture, and varying the training set. Varying the initial weights, however, requires a full optimization for each starting point and thus consumes too many computational resources. For the same reason, varying the network architecture demands caution when it is not possible to share learned information between different architectures. Unlike previous possibilities, varying the training set can be done in any point of the optimization process for one or multiple network architectures, and, depending on how it is done, it may not impact the final training cost.

\begin{figure}[!htb]
 \centering
 \begin{tikzpicture}
   \tikzset{blockr/.style={draw, rectangle, text centered, drop shadow, fill=white, text width=1.37cm, minimum height=0.55cm}}
    \tikzset{blockc/.style={draw, circle, fill=black, inner sep=1.5pt}}
     \tikzset{blockd/.style={draw, circle, color=red, fill=red, inner sep=1.5pt}}
    
   \draw(3.0,10.4) node[text centered, text width=2.2cm] (text2) {
      {\normalsize Training Set}
   }; 
    
   \draw(+6.2,9.05) node[inner sep=-1pt] (image_end) {
     \includegraphics[width=2.9cm]{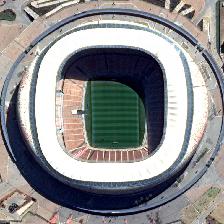}   
   };
   
   \node[inner sep=2pt, rotate=25] (dots) at (4.3,8.85) {\Large \ldots}; 
       
   \draw(+2.4,8.75) node[inner sep=-1pt] (image2) {
     \includegraphics[width=2.9cm]{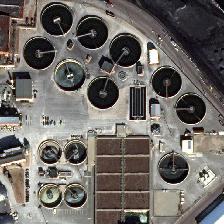}   
   };
   
   \draw(+1.4,8.6) node[inner sep=-1pt] (image2) {
     \includegraphics[width=2.9cm]{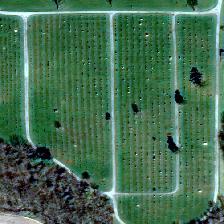}   
   };
   
   \draw(+0.4,8.45) node[inner sep=-1pt] (image) {
      \includegraphics[width=2.9cm]{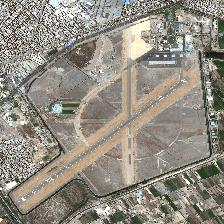}   
   };
      
   \path[->](1.8,4.7) node[blockr] (r1) {
      \footnotesize ResNet
   };
   
   
   \path[->](4.1,4.7) node[blockr] (wr) {
      \footnotesize Body weights
   };
   
   \path[->](6.7,5.7) node[blockr] (r3) {
      \footnotesize ResNet$_1$
   };
   
   \path[->](6.7,5.0) node[blockr] (r4) {
      \footnotesize ResNet$_2$
   };
   
   \path[->](6.7,4.3) node[blockr] (r5) {
      \footnotesize ResNet$_3$
   };
   
   \draw(6.7,3.8) node[text centered, text width=1.8cm] (l1) {
      {\large $\dots$}
   }; 

   \path[->](6.7,3.4) node[blockr] (r6) {
      \footnotesize ResNet$_m$
   };

   \draw [->,thick](r1)--(wr);
   \draw (wr) edge[out=0,in=180,->] (r3);
   \draw (wr) edge[out=0,in=180,->] (r4);
   \draw (wr) edge[out=0,in=180,->] (r5);
   \draw (wr) edge[out=0,in=180,->] (r6);
   \draw (wr) edge[out=0,in=180,->] (l1);
   
   \draw[thick,draw=black,dashed] (0.8,+1.1) -- (2.8,+1.1) -- (2.8,5.15) -- (0.8,5.15) -- cycle;
      
   \draw[thick,draw=black,dashed] (5.7,-0.2) -- (7.7,-0.2) -- (7.7,6.2) -- (5.7,6.2) -- cycle;   
   
   \draw[draw=black,thick] (-1.2,6.9) -- (7.8,6.9) -- (7.8,10.7) -- (-1.2,10.7) -- cycle; 
      
    
   \draw(1.8,+0.7) node[text centered, text width=2.2cm] (text2) {
      {\small Hydra's body}
   };
   
   \draw(6.7,-0.6) node[text centered, text width=2.2cm] (text2) {
      {\small Hydra's heads}
   };
   
   \path[->](1.8,1.6) node[blockr] (d1) {
      \footnotesize DenseNet
   };
   
   \path[->](4.1,1.6) node[blockr] (wd) {
      \footnotesize Body weights
   };
      
   \path[->](6.7,2.6) node[blockr] (d2) {
      \footnotesize DenseNet$_1$
   };
   
   \path[->](6.7,1.9) node[blockr] (d3) {
      \footnotesize DenseNet$_2$
   };
   
   \path[->](6.7,1.2) node[blockr] (d4) {
      \footnotesize DenseNet$_3$
   };
   
   \draw(6.7,0.75) node[text centered, text width=1.8cm] (l2) {
      {\large $\dots$}
   }; 

   \path[->](6.7,0.3) node[blockr] (d6) {
      \footnotesize DenseNet$_n$
   };
   
   \draw (d1) edge[out=0,in=180,->] (wd);
   \draw (wd) edge[out=0,in=180,->] (d2);
   \draw (wd) edge[out=0,in=180,->] (d3);
   \draw (wd) edge[out=0,in=180,->] (d4);
   \draw (wd) edge[out=0,in=180,->] (l2);
   \draw (wd) edge[out=0,in=180,->] (d6);

   
   \path[->](-0.5,3.1) node[blockr] (imgnet) {
      \footnotesize ImageNet weights
   };
   
   \draw (imgnet) edge[out=0,in=180,->] (d1);
   \draw (imgnet) edge[out=0,in=180,->] (r1);


   
   \draw (6.6,6.9) edge[->] (6.6,6.3);
   \draw (1.8,6.9) edge[->] (1.8,5.3);
   
   
 
\end{tikzpicture} 
  \caption{\revision{Hydra's training flowchart: two CNN architectures, ResNet and DenseNet, are initialized with ImageNet weights and coarsely optimized to form the Hydra's body. The obtained weights are then refined several times through further optimization to create the heads of the Hydra. ResNets and DesNets use the same training set in both optimization stages (Hydra's body and Hydra's heads).}} 
  \label{fig:hydra_ensemble}
\end{figure}
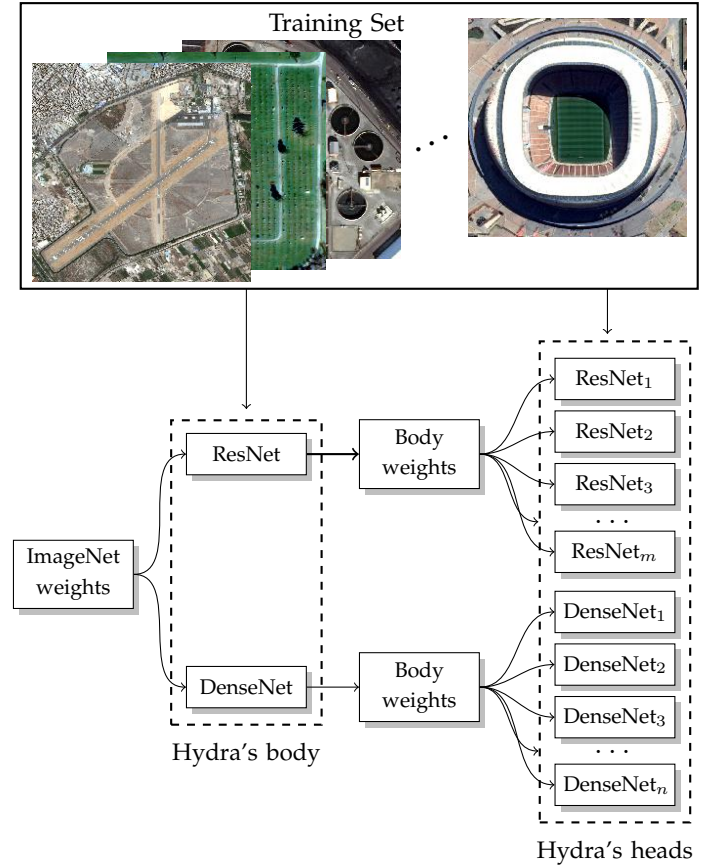

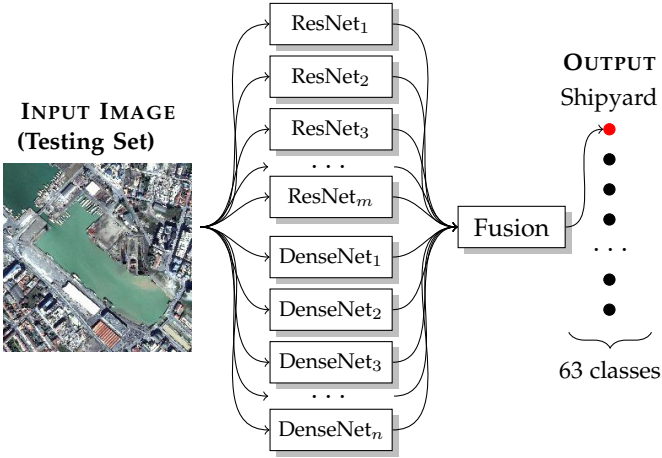
\begin{figure}[!htb]
 \centering
 \begin{tikzpicture}
   \tikzset{blockr/.style={draw, rectangle, text centered, drop shadow, fill=white, text width=1.4cm, minimum height=0.55cm}}
   \tikzset{blockt/.style={draw, rectangle, text centered, drop shadow, fill=white, text width=1.2cm, minimum height=0.55cm}}
    \tikzset{blockc/.style={draw, circle, fill=black, inner sep=1.5pt}}
     \tikzset{blockd/.style={draw, circle, color=red, fill=red, inner sep=1.5pt}}
    
   \draw(+2.9,3.0) node[text centered, text width=2.7cm, inner sep=0pt] (image) {
      {\small \textbf{\textsc{Input Image} (Testing Set)}$\quad$}\\
      \vspace{0.1cm}
      \includegraphics[width=2.5cm]{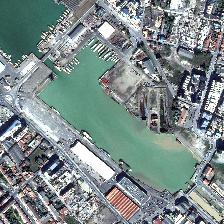}   
   };
    
   \path[->](6.0,5.7) node[blockr] (r3) {
      \footnotesize ResNet$_1$
   };
   
   \path[->](6.0,5.0) node[blockr] (r4) {
      \footnotesize ResNet$_2$
   };
   
   \path[->](6.0,4.3) node[blockr] (r5) {
      \footnotesize ResNet$_3$
   };
   
   \draw(6.0,3.8) node[text centered, text width=1.8cm, inner sep=-2pt] (l1) {
      {\large $\dots$}
   }; 

   \path[->](6.0,3.4) node[blockr] (r6) {
      \footnotesize ResNet$_m$
   };

   \draw (image) edge[out=0,in=180,->] (r3);
   \draw (image) edge[out=0,in=180,->] (r4);
   \draw (image) edge[out=0,in=180,->] (r5);
   \draw (image) edge[out=0,in=180,->] (r6);
   \draw (image) edge[out=0,in=180,->] (l1);
         
   \path[->](6.0,2.6) node[blockr] (d2) {
      \footnotesize DenseNet$_1$
   };
   
   \path[->](6.0,1.9) node[blockr] (d3) {
      \footnotesize DenseNet$_2$
   };
   
   \path[->](6.0,1.2) node[blockr] (d4) {
      \footnotesize DenseNet$_3$
   };
   
   \draw(6.0,0.75) node[text centered, text width=1.8cm, inner sep=-2pt] (l2) {
      {\large $\dots$}
   }; 

   \path[->](6.0,0.3) node[blockr] (d6) {
      \footnotesize DenseNet$_n$
   };
   
   \draw (image) edge[out=0,in=180,->] (d2);
   \draw (image) edge[out=0,in=180,->] (d3);
   \draw (image) edge[out=0,in=180,->] (d4);
   \draw (image) edge[out=0,in=180,->] (l2);
   \draw (image) edge[out=0,in=180,->] (d6);
   
   \path[->](8.4,3.0) node[blockt] (e0) {
      \normalsize Fusion
   };
   
   \draw (r3) edge[out=0,in=180,->] (e0);
   \draw (r4) edge[out=0,in=180,->] (e0);
   \draw (r5) edge[out=0,in=180,->] (e0);
   \draw (r6) edge[out=0,in=180,->] (e0);
   \draw (l1) edge[out=0,in=180,->] (e0);
   
   \draw (d2) edge[out=0,in=180,->] (e0);
   \draw (d3) edge[out=0,in=180,->] (e0);
   \draw (d4) edge[out=0,in=180,->] (e0);
   \draw (l2) edge[out=0,in=180,->] (e0);
   \draw (d6) edge[out=0,in=180,->] (e0);
   
   \path[->](9.7,4.3) node[blockd] (c0) {};
   \path[->](9.7,3.9) node[blockc] (c1) {};
   \path[->](9.7,3.5) node[blockc] (c2) {};
   \path[->](9.7,3.1) node[blockc] (c3) {};
   
   \draw(9.75,2.7) node[text centered, text width=1.8cm] (dots2) {
      {\large $\dots$}
   };
   
   \path[->](9.7,2.3) node[blockc] (c4) {};
   \path[->](9.7,1.9) node[blockc] (c5) {};
   
   \draw (e0) edge[out=0,in=180,->] (c0);
   
   \draw(9.7,4.7) node[text centered, text width=1.8cm] (text3) {
      {\small Shipyard}
   };
   
   \draw(9.7,5.2) node[text centered, text width=1.8cm] (text4) {
      {\small \textbf{\textsc{Output}}}
   };
     
   \draw [decorate,decoration={brace,amplitude=6pt,mirror},xshift=-4pt,yshift=0pt] (9.3,1.6) -- (10.4,1.6) node [black,midway,xshift=+0.0cm, yshift=-0.5cm] 
{\small 63 classes};


\end{tikzpicture} 
  \caption{Hydra's inference flowchart: an input image is fed to all heads of the Hydra, and each head outputs the probability of each class being the correct one. A majority voting is then used to determine the final label, considering that each head votes for its most probable label.}
  \label{fig:hydra_inference}
\end{figure}

\revision{Our framework uses two state-of-the-art CNN architectures, Residual Networks (ResNet)~\cite{He2015} and Dense Convolutional Networks (DenseNet)~\cite{huang2017densely}, as shown in Figures~\ref{fig:hydra_ensemble}~and~\ref{fig:hydra_inference}. They were chosen after experimenting with them and other architectures: VGG-16~\cite{Simonyan2015}, VGG-19~\cite{Simonyan2015}, Inception V3~\cite{Szegedy2015}, and Xception~\cite{Chollet2016}. Different reasons contribute into making ResNet and DenseNet the best combination of architectures: (1) they outperform any other evaluated architecture and other architectures previously mentioned in Section~\ref{sec:related} (see Table~\ref{table:nwpu-resisc45_performance}); (2) they have a reasonable level of complementarity; and (3) they achieve similar classification results individually.} The training process depicted in Figure~\ref{fig:hydra_ensemble} consists in first creating the body of the Hydra by coarsely optimizing the chosen architectures using weights pre-trained on \textsc{ImageNet}~\cite{deng2009} as a starting point. Then, the obtained weights serve as a starting point for several copies of these CNNs, which are further optimized to form the heads of the Hydra. During testing, an image is fed to all Hydra's heads and their outputs are fused to generate a single decision, as illustrated in Figure~\ref{fig:hydra_inference}.

A key point on ensemble learning, as observed by Krogh~\etal~\cite{Krogh1994} and Xandra~\etal~\cite{Chandra2006}, is that the hypotheses should be as accurate and as diverse as possible. In Hydra, such properties were prompted by applying different geometric transformations over the training samples, like zoom, shift, reflection, and rotations. More details regarding the proposed framework are presented in Sections~\ref{sec:cnns},~\ref{sec:data_augmentation}~and~\ref{sec:fusion}.

\subsection{CNN architectures}
\label{sec:cnns}

Both CNN architectures used in this work, ResNet (ResNet-50)~\cite{He2015} and DenseNet (DenseNet-161)~\cite{huang2017densely}, were chosen for achieving state-of-the-art results in different classification problems while being complementary to each other. Let $\mathcal{F}_{\ell}(\cdot)$ a non-linear transformation --- a composite function of convolution, relu, pooling, batch normalization and so on --- for layer $\ell$. As described by He~\etal~\cite{He2015}, a ResNet unit can be defined as:

\begin{equation}
   x_{\ell} = \mathcal{F}_{\ell} (x_{\ell-1}) + x_{\ell-1}
   \label{eq:resnet} 
\end{equation}

\noindent where $x_{\ell-1}$ and $x_{\ell}$ are the input and output of layer $\ell$, and $x_0$ is the input image. ResNets are built by stacking several of these units, creating bypasses for gradients that improve the back-propagation optimization in very deep networks. As a consequence, ResNets may end up generating redundant layers. As a different solution to improve the flow of gradients in deep networks, Huang~\etal~\cite{huang2017densely} proposed to connect each layer to all preceding layers, which was named a DenseNet unit:

\begin{equation}
   x_{\ell} = \mathcal{F}_{\ell}([x_0, x_1, \dots, x_{\ell-1}])
\end{equation}

\noindent where $[x_0, x_1, \dots, x_{\ell-1}]$ is the concatenation of the $\ell$  previous layer outputs. DenseNets also stack several units, but they use transition layers to limit the number of connections. In the end, DenseNets have less parameters than ResNets and compensate this facts through feature reuse. This difference in behavior may be one of the causes for their complementarity.

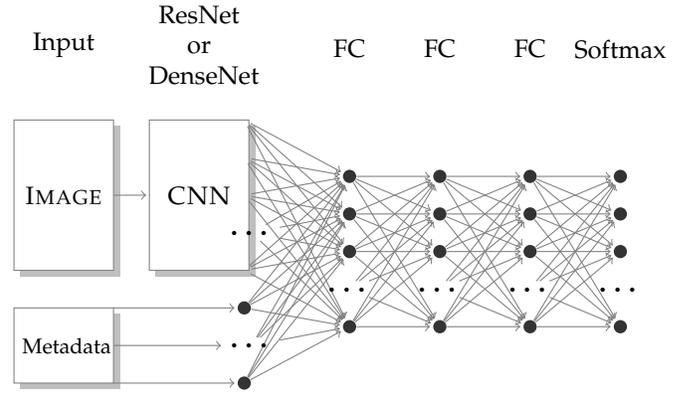
\begin{figure}[!htb]
 \centering
 \def\layersep{1.2cm}
 \begin{tikzpicture}[shorten >=1pt,->,draw=black!50, node distance=\layersep]
    \tikzstyle{every pin edge}=[<-,shorten <=1pt]
    \tikzstyle{neuron}=[circle,fill=black!25,minimum size=5pt,inner sep=0pt]
     \tikzstyle{white neuron}=[neuron, fill=white, color=white];
    \tikzstyle{input neuron}=[neuron, fill=black!80];
    \tikzstyle{output neuron}=[neuron, fill=black!80];
    \tikzstyle{hidden neuron}=[neuron, fill=black!80];
    \tikzstyle{annot} = [text width=4em, text centered]

	\tikzset{blockr/.style={draw, rectangle, text centered, drop shadow, fill=white, text width=1.1cm, minimum height=2.0cm}}

	\tikzset{blockr2/.style={draw, rectangle, text centered, drop shadow, fill=white, text width=1.1cm, minimum height=1.0cm}}

   \foreach \name / \y in {1/1.0,2/1.5,3/2.0,4/3.0}
       \node[white neuron] (WHITE-\name) at (-1.4,-\y) {};
   \foreach \name / \y in {5/3.5,6/4.5}
       \node[input neuron] (WHITE-\name) at (-1.4,-\y) {};

   \path[->](-3.8,-2.0) node[blockr] (img) {
      \normalsize \textsc{Image}
   };

   \path[->](-3.8,-4.0) node[blockr2] (meta) {
      \footnotesize Metadata
   };

   \path[->](-2.0,-2.0) node[blockr] (cnn) {
      \normalsize \textsc{CNN}
   };

    \foreach \name / \y in {1/1.0,2/1.5,3/2.0,4/3.0}
       \node[input neuron] (I-\name) at (0,-\y-0.75) {};

    \foreach \name / \y in {1/1.0,2/1.5,3/2.0,4/3.0}
       \node[input neuron] (HA\name) at (\layersep,-\y-0.75) {};

    \foreach \name / \y in {1/1.0,2/1.5,3/2.0,4/3.0}
        \path[yshift=0.0cm]
            node[hidden neuron] (HB\name) at (2*\layersep,-\y-0.75) {};

    \foreach \name / \y in {1/1.0,2/1.5,3/2.0,4/3.0}
        \node[output neuron] (O-\name) at (3*\layersep,-\y-0.75) {};

    \foreach \source in {1,...,4}
        \foreach \dest in {1,...,4}
            \path (I-\source) edge (HA\dest);
            
    \foreach \source in {1,...,6}
        \foreach \dest in {1,...,4}
            \path (WHITE-\source) edge (I-\dest);

    \node[inner sep=2pt] (dots0) at (-1.3,-2.5) {\Large \ldots};
               
    \foreach \dest in {1,...,4}
       \path (dots0) edge (I-\dest);        

    \node[inner sep=2pt] (dots00) at (-1.3,-4.0) {\Large \ldots};

    \foreach \dest in {1,...,4}
       \path (dots00) edge (I-\dest);     

    \node[inner sep=2pt] (dots1) at (0.0,-3.25) {\Large \ldots};
               
    \foreach \dest in {1,...,4}
       \path (dots1) edge (HA\dest);
       
    \node[inner sep=2pt] (dots2) at (\layersep,-3.25) {\Large \ldots};
    
    \foreach \dest in {1,...,4}
       \path (dots2) edge (HB\dest);
       
    \node[inner sep=2pt] (dots3) at (2*\layersep,-3.25) {\Large \ldots};
    
    \foreach \dest in {1,...,4}
       \path (dots3) edge (O-\dest);
       
    \node[inner sep=2pt] (dots4) at (3*\layersep,-3.25) {\Large \ldots};

    \foreach \source in {1,...,4}
        \foreach \dest in {1,...,4}
            \path (HA\source) edge (HB\dest);
            
    \foreach \source in {1,...,4}
        \foreach \dest in {1,...,4}
            \path (HB\source) edge (O-\dest);

    \draw[->] (img.east) |- (cnn.west);
    \draw[->] (meta.north east) |- (WHITE-5.west);
    \draw[->] (meta.south east) |- (WHITE-6.west);
    \draw[->] (meta.east) |- (dots00.west);
    
    \node[annot,above of=img, node distance=2cm] (hl) {Input};
    \node[annot,above of=cnn, node distance=2cm] (hl) {ResNet \\or\\ DenseNet};
    \node[annot,above of=HA1, node distance=1.70cm] (hl) {FC};
    \node[annot,above of=I-1, node distance=1.70cm] {FC};
    \node[annot,above of=HB1, node distance=1.70cm] (dense3) {FC};
    \node[annot,above of=O-1, node distance=1.70cm] {Softmax};
\end{tikzpicture}
 
  \caption{The convolutional neural network architecture used by the Hydra framework.} 
  \label{fig:hydra_block}
\end{figure}

\revision{We initialize both DenseNet and ResNet with weights from ImageNet~\cite{deng2009}. For each of them, the last fully connected layer ({\it i.e.}, the classification layer) is replaced by three hidden layers with 4096 neurons and a new classification layer, as illustrated in Figure~\ref{fig:hydra_block}. The extension in these architectures is necessary to take image metadata into account. A vector of metadata that is useful for classification ({\it e.g.}, ground sample distance, sun angle) is concatenated to the input of the first hidden layer so that subsequent layers learn the best way to combine metadata and visual features.}


\subsection{On-line data augmentation}
\label{sec:data_augmentation}

The process of artificially augmenting the image samples in machine learning by using class-preserving transformations is meant to reduce the overfitting in imbalanced classification problems or to improve the generalization ability in the absence of enough data~\cite{Krizhevsky2017,DBLP:journals/corr/Howard13,7005506,43022,7299016}. The breakthroughs achieved by many state-of-the-art algorithms that benefit from data augmentation lead many authors~\cite{ChatfieldSVZ1,2017arXiv170205538D,2017arXiv171204621P,7797091} to study the effectiveness of this technique.

Different techniques can be considered, such as geometric ({\it e.g.}, reflection, scaling, warping, rotation and translation) and photometric ({\it e.g.}, brightness and saturation enhancements, noise perturbation, edge enhancement, and color palette changing) transformations. However, as pointed out by Ratner~\etal~\cite{Ratner2017}, selecting transformations and tuning their parameters can be a tricky and time-consuming task. A careless augmentation may produce unrealistic or meaningless images that will negatively affect the performance.

\begin{figure}[!htb]

 \begin{tikzpicture}
   \tikzset{blockr/.style={draw, rectangle, text centered, drop shadow, fill=white, text width=1.8cm, minimum height=0.55cm}}
    \tikzset{blockc/.style={draw, circle, fill=black, inner sep=1.5pt}}
     \tikzset{blockd/.style={draw, circle, color=red, fill=red, inner sep=1.5pt}}
    
    
   \draw [decorate,decoration={brace,amplitude=8pt},xshift=+4pt,yshift=0pt] (-1.2,9.7) -- (6.5,9.7) node [black,midway,xshift=+0.0cm, yshift=+0.5cm] 
{\small Training samples ($n$ regions)};
    
   \draw(5.3,8.50) node[text centered, text width=3.8cm, inner sep=0pt] (img1) {
      \includegraphics[width=2.4cm]{figures/airport_13_3.jpg}   
   };
   
   \draw(2.0,8.00) node[text centered, text width=3.8cm, inner sep=0pt] (img1) {
      \includegraphics[width=2.4cm]{figures/water_treatment_facility_104_0.jpg}   
   };
   
   \draw(1.0,7.75) node[text centered, text width=3.8cm, inner sep=0pt] (img1) {
      \includegraphics[width=2.4cm]{figures/burial_site_870_0.jpg}   
   }; 
    
   \node[inner sep=2pt, rotate=25] (dots) at (3.7,8.00) {\Large \ldots}; 
    
   \draw(0.0,7.50) node[text centered, inner sep=15pt] (fountain) {
      \includegraphics[width=2.4cm]{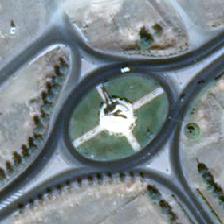}   
   };

   \draw(3.00,5.9) node[text centered, text width=5.0cm] (text2) {
      {Off-line data augmentation}
   };
   
   
   \draw(-1.00,4.8) node[text centered, inner sep=0pt, rotate=270] (aug1) {
      \includegraphics[width=1.2cm]{figures/fountain_630_0.jpg}   
   };
   
   \draw(+0.25,4.8) node[text centered, inner sep=0pt, rotate=180] (aug2) {
      \includegraphics[width=1.2cm]{figures/fountain_630_0.jpg}   
   };
   
   \draw(+1.50,4.8) node[text centered, inner sep=0pt, rotate=90] (aug3) {
      \includegraphics[width=1.2cm]{figures/fountain_630_0.jpg}   
   };
   
   \draw(+2.75,4.8) node[text centered, inner sep=0pt, rotate=0] (aug4) {
      \includegraphics[width=1.2cm]{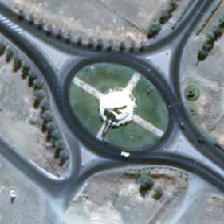}   
   };
   
   \draw(+4.00,4.8) node[text centered, inner sep=0pt, rotate=90] (aug5) {
      \includegraphics[width=1.2cm]{figures/fountain_630_0_flip.jpg}   
   };
   
   \draw(+5.25,4.8) node[text centered, inner sep=0pt, rotate=180] (aug6) {
      \includegraphics[width=1.2cm]{figures/fountain_630_0_flip.jpg}   
   };
   
   \draw(+6.50,4.8) node[text centered, inner sep=0pt, rotate=270] (aug7) {
      \includegraphics[width=1.2cm]{figures/fountain_630_0_flip.jpg}   
   };
   
   
   \draw[->] (fountain.south) -| (aug1);
   \draw[->] (fountain.south) -| (aug2);
   \draw[->] (fountain.south) -| (aug3);
   \draw[->] (fountain.south) -| (aug4);
   \draw[->] (fountain.south) -| (aug5);
   \draw[->] (fountain.south) -| (aug6);
   \draw[->] (fountain.south) -| (aug7);
  
    
   \draw [decorate,decoration={brace,amplitude=8pt},xshift=+4pt,yshift=0pt] (-1.2,2.2) -- (6.5,2.2) node [black,midway,xshift=+0.0cm, yshift=+0.5cm] 
{\small Image samples for all epochs ($8n$ regions)};
    
   \draw(5.3,1.00) node[text centered, inner sep=0pt, rotate=180] (img1) {
      \includegraphics[width=2.4cm]{figures/airport_13_3.jpg}   
   };
   
   \draw(2.0,0.50) node[text centered, inner sep=0pt, rotate=180] (img1) {
      \includegraphics[width=2.4cm]{figures/fountain_630_0.jpg}   
   };
   
   \draw(1.0,0.25) node[text centered, inner sep=0pt, rotate=270] (img1) {
      \includegraphics[width=2.4cm]{figures/fountain_630_0.jpg}   
   }; 
    
   \node[inner sep=2pt, rotate=25] (dots) at (3.7,0.50) {\Large \ldots}; 
    
   \draw(0.0,0.00) node[text centered, inner sep=0pt] (img1) {
      \includegraphics[width=2.4cm]{figures/fountain_630_0.jpg}   
   };
   
   \draw[-] (0.0,6.35) -| (0.0,5.8);

   \draw[->] (2.75,3.9) -| (2.75,3.1);
   \draw(4.0,3.5) node[text centered, text width=5.0cm]  {
      {CNN-Training}
   };
   
\end{tikzpicture}
 
  \caption{Illustration of an off-line data augmentation process. In this example considering basic rotations (90, 180 and 270 degrees) and reflections only, each image is augmented to a set of 8 images before the training starts. After that, the augmented training set is static for all CNN epochs.} 
  \label{fig:offline_data_augmentation}
\end{figure}

The process of data augmentation can be conducted off-line, as illustrated in Figure~\ref{fig:offline_data_augmentation}. In this case, training images undergo geometric and/or photometric transformations before the training. Although this approach avoids repeating the same transformation several times, it also considerably increases the size of the training set. For instance, if we only consider basic rotations (90, 180 and 270 degrees) and reflections, the FMOW training set would have more than 2.8 million samples. This would not only increase the amount of required storage space, but the time consumed by a training epoch as well.

\begin{figure}[!htb]
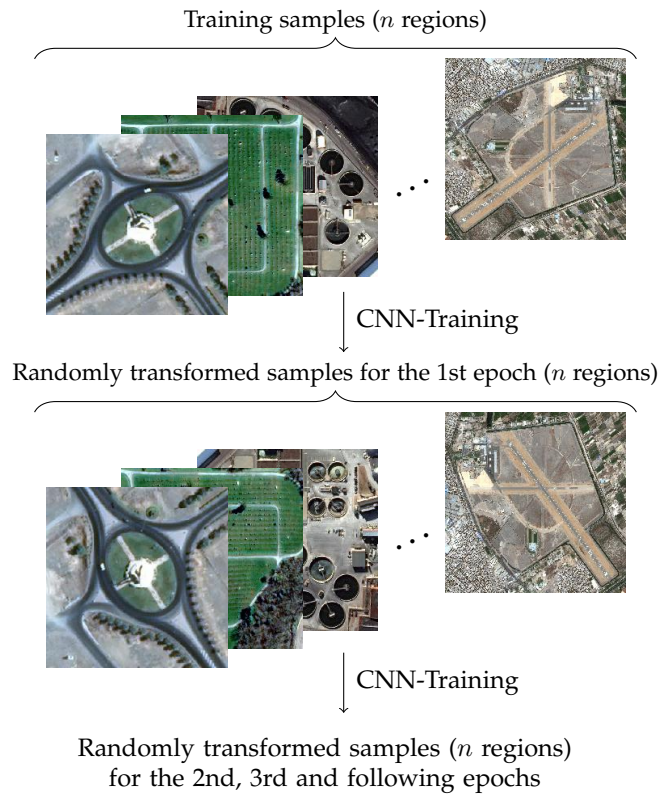

 \centering
 \begin{tikzpicture}
   \tikzset{blockr/.style={draw, rectangle, text centered, drop shadow, fill=white, text width=1.8cm, minimum height=0.55cm}}
    \tikzset{blockc/.style={draw, circle, fill=black, inner sep=1.5pt}}
     \tikzset{blockd/.style={draw, circle, color=red, fill=red, inner sep=1.5pt}}
    
    
   \draw [decorate,decoration={brace,amplitude=8pt},xshift=+4pt,yshift=0pt] (-1.5,6.9) -- (6.5,6.9) node [black,midway,xshift=+0.0cm, yshift=+0.5cm] 
{\small Training samples ($n$ regions)};
    
   \draw(5.3,5.70) node[text centered, text width=3.8cm, inner sep=0pt] (img1) {
      \includegraphics[width=2.4cm]{figures/airport_13_3.jpg}   
   };
   
   \draw(2.0,5.20) node[text centered, text width=3.8cm, inner sep=0pt] (img1) {
      \includegraphics[width=2.4cm]{figures/water_treatment_facility_104_0.jpg}   
   };
   
   \draw(1.0,4.95) node[text centered, text width=3.8cm, inner sep=0pt] (img1) {
      \includegraphics[width=2.4cm]{figures/burial_site_870_0.jpg}   
   }; 
    
   \node[inner sep=2pt, rotate=25] (dots) at (3.7,5.20) {\Large \ldots}; 
    
   \draw(0.0,4.70) node[text centered, text width=3.8cm, inner sep=0pt] (img1) {
      \includegraphics[width=2.4cm]{figures/fountain_630_0.jpg}   
   };
   
   \draw[->] (2.75,3.8) -| (2.75,3.0);
   \draw(4.0,3.4) node[text centered, text width=5.0cm]  {
      {CNN-Training}
   };
   
    
   \draw [decorate,decoration={brace,amplitude=8pt},xshift=+4pt,yshift=0pt] (-1.5,2.2) -- (6.5,2.2) node [black,midway,xshift=+0.0cm, yshift=+0.5cm] 
{\small Randomly transformed samples for the 1st epoch ($n$ regions)};
    
   \draw(5.3,1.00) node[text centered, inner sep=0pt, rotate=90] (img1) {
      \includegraphics[width=2.4cm]{figures/airport_13_3.jpg}   
   };
   
   \draw(2.0,0.50) node[text centered, inner sep=0pt, rotate=180] (img1) {
      \includegraphics[width=2.4cm]{figures/water_treatment_facility_104_0.jpg}   
   };
   
   \draw(1.0,0.25) node[text centered, inner sep=0pt, rotate=90] (img1) {
      \includegraphics[width=2.4cm]{figures/burial_site_870_0.jpg}   
   }; 
    
   \node[inner sep=2pt, rotate=25] (dots) at (3.7,0.50) {\Large \ldots}; 
    
   \draw(0.0,0.00) node[text centered, inner sep=0pt, rotate=90] (img1) {
      \includegraphics[width=2.4cm]{figures/fountain_630_0.jpg}   
   };
   
   \draw(2.5,-2.5) node[text centered, text width=8.0cm]  {
      {Randomly transformed samples ($n$~regions) for the 2nd, 3rd and following epochs}
   };
   
   \draw[->] (2.75,-1.0) -| (2.75,-1.8);
   \draw(4.0,-1.4) node[text centered, text width=5.0cm]  {
      {CNN-Training}
   };
   
\end{tikzpicture}
 
  \caption{Illustration of an on-line data augmentation process. In this example considering basic rotations (90, 180 and 270 degrees) and reflections only, each image is randomly transformed into a new image before each training epoch starts.} 
  \label{fig:online_data_augmentation}
\end{figure}

In our framework, we apply an on-line augmentation, as illustrated in Figure~\ref{fig:online_data_augmentation}. To this end, different random transformations were applied to the training samples every epoch. This way, the number of transformations seen during training increase with the number of epochs. In addition, different heads of the Hydra experience different versions of the training set and hardly converge to the same endpoint.

On-line data augmentations were carried out by the ImageDataGenerator function from Keras\footnote{\url{https://keras.io/}}. We used random flips in vertical and horizontal directions, random zooming and random shifts over different image crops. We did not use any photometric distortion to avoid creating meaningless training samples.

\subsection{Fusion and decision}
\label{sec:fusion}

\revision{To discover the most appropriate label for an input region according to the models created by Hydra, we have to combine the results from all heads. Each head produces a score vector that indicates how well this region matches each class. If a region shows up in more than one image, it will have multiple score vectors per head. In this case, these vectors are summed to form a single vector of scores per head for a region. After that, each head votes for the class with the highest score within its score vector for that region. Finally, a majority voting is used to select the final label. But if the number of votes is below or equal to half the number of heads, the region is considered a false detection.}

\section{Experimental results}~\label{sec:experiments}

\subsection{The FMOW dataset}
\label{sec:challenge}

The main goal of the FMOW challenge was to encourage researchers and machine learning experts to design automatic classification solutions for land use interpretation in satellite images. The competition was hosted by TopCoder\footnote{\url{https://www.topcoder.com/}}, where participants had one submission every three hours to keep track of their ongoing performance. Details regarding dataset, scoring, and restrictions are given in Sections~\ref{sec:fmow_dataset},~\ref{sec:fmow_scoring},~and~\ref{sec:fmow_restriction}, respectively. Training details for FMOW are presented in Section~\ref{sec:fmow_training} and our results in  Section~\ref{sec:fmow_results}.

\subsubsection{Dataset}
\label{sec:fmow_dataset}

The FMOW dataset, as detailed by Christie~\emph{et al.}~\cite{fmow2017}, contains almost half of a million images split into training, evaluation and testing subsets, as presented in Table~\ref{tab:fmow_dataset}. It was publicly released\footnote{\url{https://www.iarpa.gov/challenges/fmow.html}} in two image formats: JPEG (194 GB) and TIFF (2.8 TB). For both formats, high-resolution pan-sharpened~\cite{li2015} and multi-spectral images were available, although the latter one varies with the format. While TIFF images provided the original 4/8-band multi-spectral information, the color space was compressed in JPG images to fit their 3-channel limitation.

\begin{table}[!ht]
 \centering
 \caption{Table of contents of the FMOW dataset.}
 \begin{tabular}{lrrr}
  \toprule
  & {\bf \# of images} & {\bf \# of boxes} & {\bf \# of distinct boxes} \\
  \midrule
  Training       & 363,572  & 363,572  &  83,412 \\
  Evaluation     &  53,041  &  63,422  &  12,006 \\
  Testing        &  53,473  &  76,988  &  16,948 \\
  \bottomrule
  \textbf{Total} & 470,086  & 503,982  & 112,366 \\
  \bottomrule
 \end{tabular}
 \label{tab:fmow_dataset}
\end{table}

\begin{figure}[!ht]
 \centering
 \subfigure[]{\label{fig:fmow_challenge_a}\includegraphics[scale=0.27]{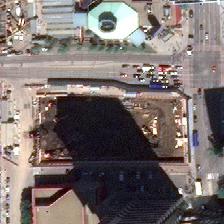} \includegraphics[scale=0.27]{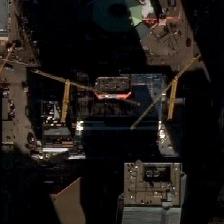} \includegraphics[scale=0.27]{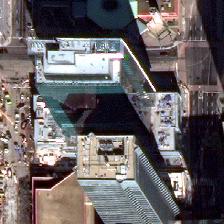} \includegraphics[scale=0.27]{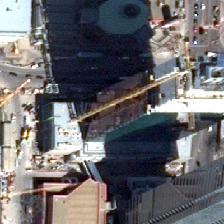}} 
 \subfigure[]{\label{fig:fmow_challenge_b}\includegraphics[scale=0.27]{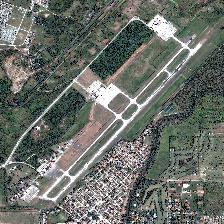} \includegraphics[scale=0.27]{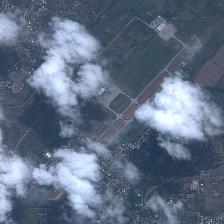}}\hspace{0.1cm}
 \subfigure[]{\label{fig:fmow_challenge_c}\includegraphics[width=0.113\textwidth]{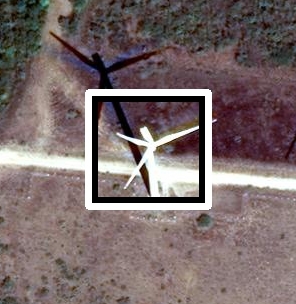} \includegraphics[width=0.113\textwidth]{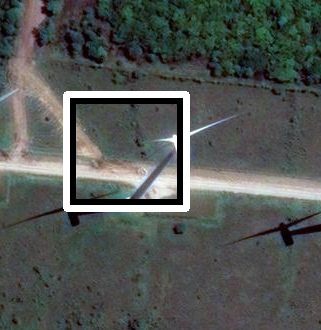}}
 \caption{Different challenges in FMOW satellite images: (a)~shadow and viewpoint variations, (b)~arbitrary weather condition and (c)~inaccurate annotations.}
 \label{fig:fmow_challenge}
\end{figure}

Pan-sharpened image dimensions range from $293\times230$ to $16184\times16288$ pixels, and multi-spectral from $74\times58$ to $4051\times4077$. Each image has one or more annotated boxes that correspond to regions of interest. Some of these regions may appear in several images taken in different time periods, adding a temporal element to the problem, as shown in Figures~\ref{fig:fmow_challenge_a},~\ref{fig:fmow_challenge_b}~and~\ref{fig:fmow_challenge_c}. Traditional problems arising from that include, but are not limited to, variations in shadows, satellite viewpoint and weather conditions. Figure~\ref{fig:fmow_challenge_c} illustrates other recurrent problems in the FMOW dataset: inaccurate annotations and large variations in region size.

\begin{figure*}[!ht]
\centering
\includegraphics[width=0.99\textwidth]{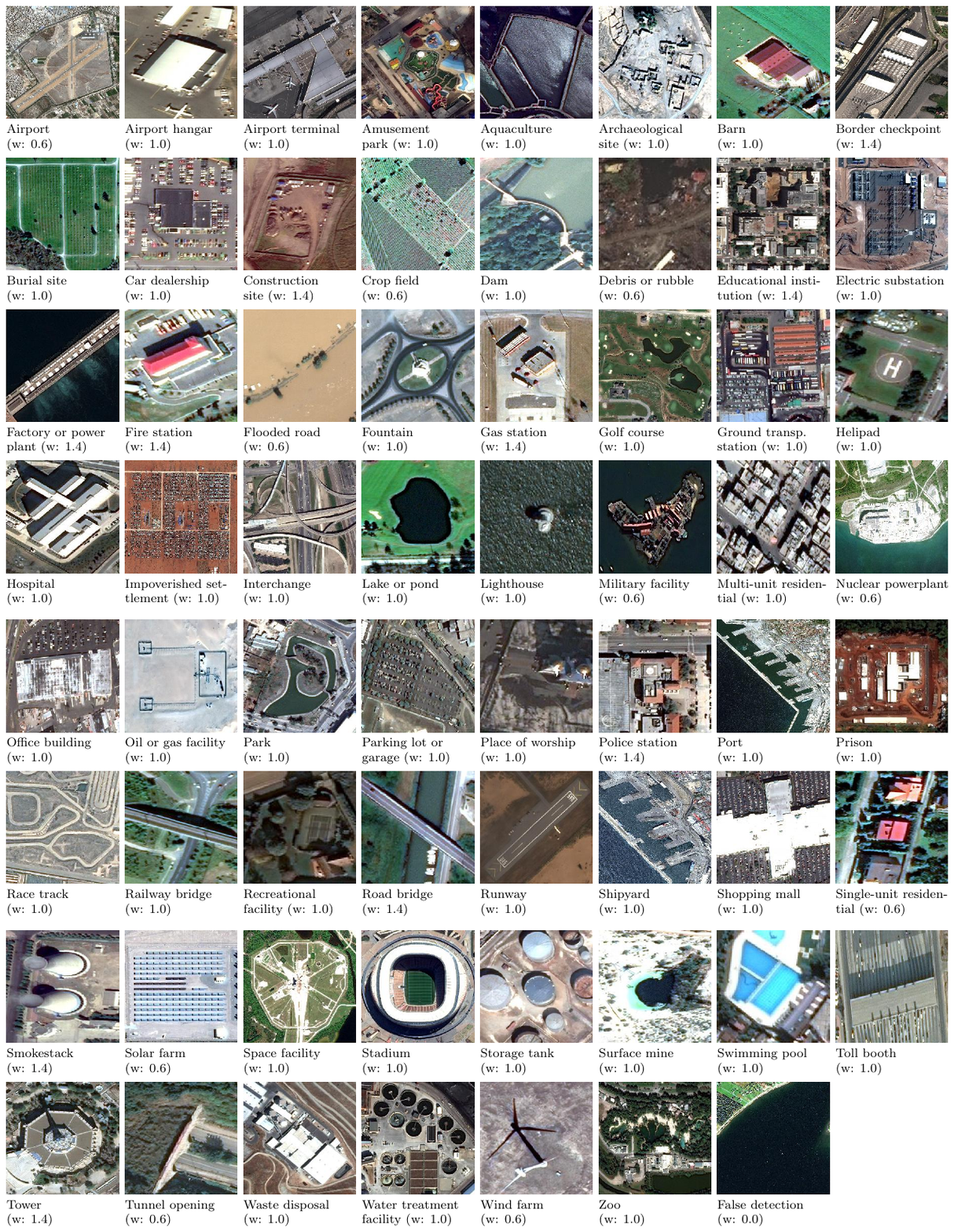}
\caption{List of classes of the FMOW challenge. An image example, a label and an associated weight are presented for each class. The least challenging classes have a lower weight ($0.6$) while the most difficult ones have a higher weight ($1.4$). These weights impact the final classification score.}
\label{fig:fmow_classes}
\end{figure*}

\begin{figure*}[!ht]
\centering
\includegraphics[scale=0.43]{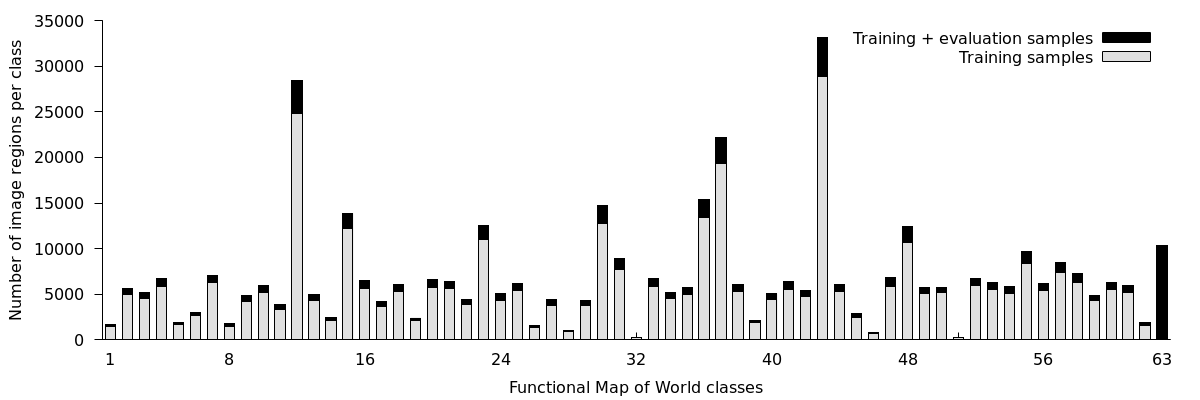}  
\caption{The class histogram distribution for the \texttt{training} (gray bars) and \texttt{evaluation} sets (black bars) by using the 3-band pansharpened JPEG/RGB set. Note the highly unbalanced class distribution.}
\label{fig:fmow_class_distribution}
\end{figure*}

Each region represents one of the 62 classes or a false detection. All classes considered in the FMOW challenge are illustrated in Figure~\ref{fig:fmow_classes}. Those classes have an unbalanced distribution, as shown in Figure~\ref{fig:fmow_class_distribution}. Images in the training subset contain a single region that is never a false detection. The evaluation subset has images with one or two regions, with at least one not being a false detection. Testing images may have several regions but their labels are unknown. From 53,473 images in the test subset, 82.60\% have a single region, 4.96\% have two regions, 5.66\% have three regions and 6.78\% have four regions or more, with a maximum number of fourteen regions per image.

\subsubsection{Scoring}
\label{sec:fmow_scoring}

Submissions consisted of a list with one label prediction --- among the $m$ possible classes ($m=63$, {\it i.e.} 62 land use classes plus false detection) --- for each distinct region in the test set. When the same region appeared in several images (temporal views), like the ones in Figures~\ref{fig:fmow_challenge_a}~and~\ref{fig:fmow_challenge_b}, the competitors were asked to merge all these information into a single label outcome. The quantitative criteria used by the FMOW challenge to rank its competitors was a weighted average of the $F$-measure~\cite{powers2011} for each class.

Let $P = \langle e_1, e_2, \dots, e_n \rangle$ be the predicted labels provided by a competitor and $G = \langle g_1, g_2, \dots, g_n \rangle$ the ground-truth labels for $n$ distinct test regions. From $P$ and $G$, one can derive a confusion matrix $C$ 
where $C_{i,i}$ is the number of correct classifications for the $i$-th class and $C_{i,j}$ the number of images from the $i$-th class that were mistakenly classified as the $j$-th class. With these values we compute the precision $P_i$ and recall $R_i$ scores for the $i$-th class as follows:

\begin{equation*}
 P_i = \frac{\displaystyle {tp}_i}{\displaystyle {tp}_i + {fp}_i} \qquad \qquad R_i = \frac{\displaystyle {tp}_i}{\displaystyle {tp}_i + {fn}_i}
\end{equation*}

\noindent where:

\begin{equation*}
{tp}_i = C_{i,i}  \quad\quad  {fp}_i = \sum_{j = 1,j \neq i}^{m} C_{j,i} \quad\quad {fn}_i = \sum_{j = 1, j \neq i}^{m} C_{i,j}
\end{equation*}

\noindent are respectively the number of true positives, false positives and false negatives for the $i$-th class. The $F$-measure for each class is then computed as the harmonic mean of precision and recall: $F_i = (2 \times P_i \times R_i)/(P_i + R_i)$. 

The final score $\bar{F}$ is the weighted sum of the $F$-measure for all classes:

\begin{equation}
  \bar{F} = \frac{\sum_{i = 1}^{m} F_i \times w_i}{\sum_{i = 1}^{m} w_i}
 \label{fmeasure}
\end{equation}

\noindent where $w_i$ is the weight of $i$-th class. The weights for each class are provided in Figure~\ref{fig:fmow_classes}. It is worth noting that the weight for false detections is $0$, but it does not mean it will not affect $\bar{F}$. If a region is misclassified as a false detection, its class will have an additional false negative. If a false detection is misclassified as another class, one more false positive will be taken into account.

\subsubsection{Hardware and time restrictions}
\label{sec:fmow_restriction}

The competition ran from September 21st to December 31st, 2017. The provisional standings were used to select the top 10 competitors as finalists. For the final round, competitors were asked to provide a dockerized version\footnote{\url{https://www.docker.com/}} of their solutions, and the final standings were determined based on the results for a separate sequestered dataset ({\it i.e.}, none of the competitors had access to these images). The dockerized solution had to cope with the following limitations:

\begin{itemize}
\item All models required by the solution should be generated during training using raw data only ({\it i.e.} preprocessed data was not allowed) within 7 days.
\item The prediction for all test images should be performed in no more than 24 hours. Prebuilt models should be provided to enable testing before training.
\item Hardware was restricted to two types of Linux AWS instances: \emph{m4.10xlarge} and \emph{g3.16xlarge}.
\end{itemize}

Instances of the type \emph{m4.10xlarge} have 40 Intel Xeon E5-2686 v4 (Broadwell) processors and 160GB of memory. Instances of the type \emph{g3.16xlarge} have 4 NVIDIA Tesla M60 GPUs, 64 similar processors and 488GB of memory.

\subsubsection{Baseline}
\label{sec:fmow_baseline}

The organizers released a baseline classification code\footnote{\url{https://github.com/fmow/baseline}} on October 14th, 2017, and updated that solution on November 17th of the same year. The most recent and accurate version of the baseline consists of extracting discriminant features from images and metadata using deep neural networks. To this end, they concatenated image features obtained by a CNN (DenseNet-161~\cite{huang2017densely}) to a preprocessed vector of metadata (mean subtraction) and fed this to a multilayer perceptron (MLP) with 2 hidden layers. A softmax loss is then used by the Adam optimizer~\cite{Kingma2014} during training. They used both training and validation subsets for training, and no augmentation was performed. More details can be found in the work of Christie~\emph{et al.}~\cite{fmow2017}. The classifiers employed in our Hydra framework are variations of this baseline code.

\subsubsection{FMOW training details}
\label{sec:fmow_training}

\revision{Each model is trained using the Adam optimization algorithm~\cite{Kingma2014} and softmax cross entropy. A 50\% dropout rate is used in the last three hidden layers (see Figure~\ref{fig:hydra_block}). The body of the Hydra is optimized with a learning rate of $10^{-4}$ for six epochs. The obtained weights are then used to initialize the heads of the Hydra, which are trained for five more epochs using progressive drops in the learning rate as follows: a learning rate of $10^{-4}$ is used for the first epoch, $10^{-5}$ for the following three epochs, and $10^{-6}$ for the last epoch.}

During training for the FMOW dataset, we added different class weights to the loss function for different heads seeking to enhance the complementarity between them. Four different weighting schemes were used: unweighted classes, FMOW class weights as presented in Figure~\ref{fig:fmow_classes} (false detection weight was set to $1$), a frequency-based weighting using the balanced heuristic~\cite{King2001}, and a manual adjustment of the latter.

\begin{figure}[!ht]
 \centering
 \subfigure[]{\label{fig:crop_a}\includegraphics[width=0.15\textwidth]{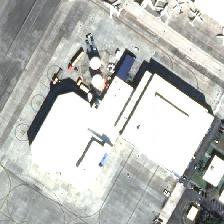}}
 \subfigure[]{\label{fig:crop_b}\includegraphics[width=0.15\textwidth]{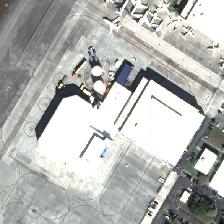}} 
 \subfigure[]{\label{fig:crop_c}\includegraphics[width=0.15\textwidth]{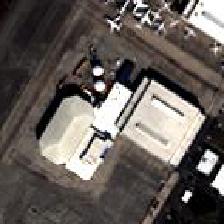}} 
 \caption{Different image crops used for training: (a)~original bounding box and (b)~expanded bounding box extracted from JPG pan-sharpened images, and (c)~expanded bounding box extracted from JPG multi-spectral images.}
 \label{fig:crops}
\end{figure}

For data augmentation, three image crop styles were considered, as illustrated in Figure~\ref{fig:crops}: the original bounding box that was provided as metadata (Figure~\ref{fig:crop_a}), and an expanded bounding box extracted from both JPG pan-sharpened images (Figure~\ref{fig:crop_b}) and JPG multi-spectral images (Figure~\ref{fig:crop_c}). As multi-spectral images have low resolution, we discarded samples with less than 96 pixels of width or height. \revision{Table~\ref{tab-heads} presents the configuration of each head in terms of network architecture, augmentation technique and class weighting, which was empirically defined after observing the performance of several combinations.}

\begin{table}[!ht]
\scriptsize
\centering
\caption{Hydra's heads description. For each head are given the network architecture, image crop style, augmentation technique and class weighting method. Three different image crop styles were considered: ORIG-PAN (Figure~\ref{fig:crop_a}), EXT-PAN (Figure~\ref{fig:crop_b}) and EXT-MULTI (Figure~\ref{fig:crop_c}). Augmentation techniques included random flips in vertical and horizontal directions (Flip), random zooming (Zoom) and random shifts (Shift). Class weighting methods were: no weighting, FMOW weights (Figure~\ref{fig:fmow_classes}) and frequency-based weighting using the balanced heuristic~\cite{King2001} before (Frequency \#2) and after (Frequency \#1) a manual adjustment.}
\label{tab-heads}
\begin{tabular}{ccccc}
\bf Head & \bf CNN & \bf Crop & \bf Augment & \bf Class weighting \\ \hline
\#1  & DenseNet & EXT-PAN   & Flip  & Unweighted    \\
\#2  & DenseNet & ORIG-PAN  & Flip  & Frequency \#1 \\
\#3  & DenseNet & EXT-MULTI & Flip  & Frequency \#1 \\
\#4  & DenseNet & EXT-PAN   & Zoom  & Frequency \#2 \\
\#5  & DenseNet & EXT-PAN   & Shift & Unweighted    \\
\#6  & DenseNet & EXT-MULTI & Shift & FMOW weights  \\
\#7  & DenseNet & EXT-PAN   & Flip  & Frequency \#2 \\
\#8  & DenseNet & ORIG-PAN  & Flip  & Frequency \#2 \\
\#9  & ResNet   & EXT-PAN   & Flip  & Unweighted    \\
\#10 & ResNet   & EXT-MULTI & Flip  & Frequency \#1 \\
\#11 & ResNet   & ORIG-PAN  & Flip  & Frequency \#1 \\
\#12 & ResNet   & EXT-MULTI & Flip  & Frequency \#2
\end{tabular}
\end{table}

\subsubsection{Results}
\label{sec:fmow_results}

In this section, we report and compare the Hydra performance in the FMOW dataset. In this experiment, we use the entire training subset and false detections of the evaluation subset for training, and the remaining images of the evaluation subset for performance computation.

\begin{figure*}[!ht]
   \centering
   \includegraphics[width=\textwidth]{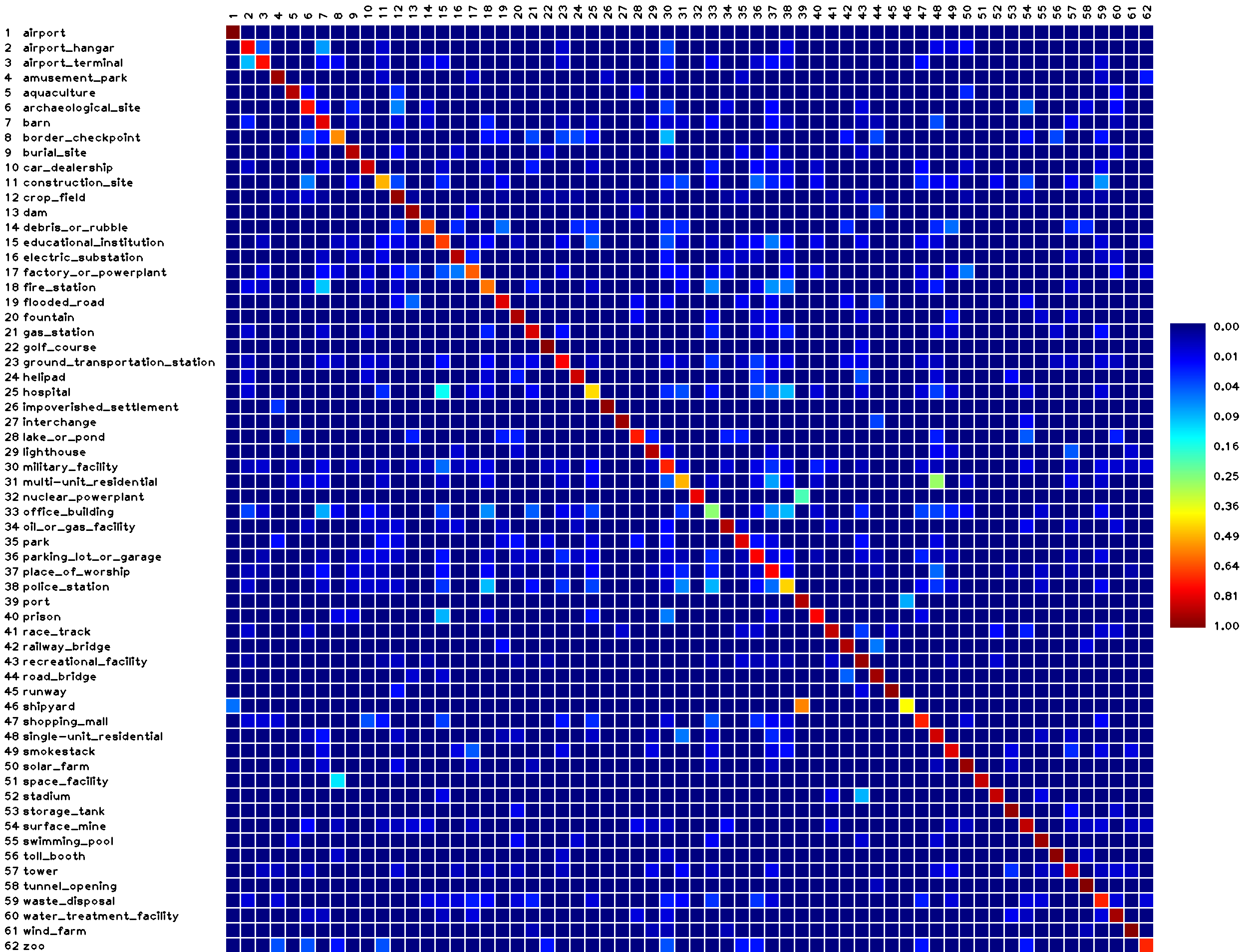}
   \caption{Confusion matrix for the Hydra ensemble after 11 training epochs (Figure~\ref{fig:ensemble_learning}). Each row represents the ground truth label, while the column shows the label obtained by the Hydra. Large values outside of the main diagonal indicate that the corresponding classes are hard to be discriminated.}    
   \label{fig:confusion_matrix_eval}
\end{figure*}

\begin{figure}[!ht]
   \subfigure[Shipyard mislabeled as port]{
      \includegraphics[width=0.21\textwidth]{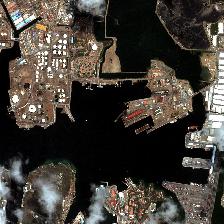}
   } \hspace{4pt}  
   \subfigure[Hospital mislabeled as educational institution.]{
      \includegraphics[width=0.21\textwidth]{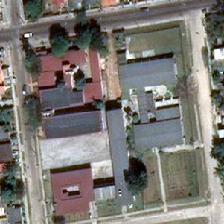}
   }\\
   \subfigure[Multi-unit residential mislabeled as single-unit residential]{
      \includegraphics[width=0.21\textwidth]{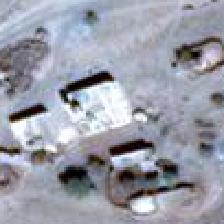}
   } \hspace{4pt}
   \subfigure[Office building mislabeled as gas station]{
      \includegraphics[width=0.21\textwidth]{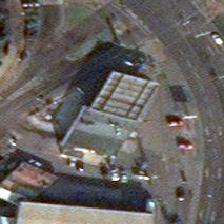}
   }
   \caption{Regions that were mislabeled by our Hydra ensemble.}    
   \label{fig:hydra_errors}
\end{figure}

The confusion matrix for the best ensemble is presented in Figure~\ref{fig:confusion_matrix_eval}. The hardest classes for Hydra were: \textbf{shipyard} (mainly confused with \textbf{port}), \textbf{hospital} (confused with \textbf{educational institution}), \textbf{multi unit residential} (confused with \textbf{single unit residential}), \textbf{police station} (confused with \textbf{educational institution}),  and \textbf{office building} (confused with \textbf{fire station}, \textbf{police station}, etc). We also had high confusion between \textbf{nuclear powerplant} and \textbf{port}, between \textbf{prison} and \textbf{educational institution}, between \textbf{space facility} and \textbf{border checkpoint}, and between \textbf{stadium} and \textbf{recreational facility}. These, however, are natural mistakes that could easily be repeated by human experts. Most of these classes look alike in aerial images ({\it e.g.}, \textbf{shipyard} and \textbf{port}) or have similar features that can trick an automatic classifier ({\it e.g.}, both \textbf{stadium} and \textbf{recreational facility} have a sports court). In Figure~\ref{fig:hydra_errors} we show examples of mislabeled regions that cannot be easily identified, even by a human.

We then reran the training process using all images from the training and evaluation subsets. The TopCoder submission system was used to obtain the $\bar{F}$-measure (Equation~\ref{fmeasure}) for the testing subset, as it has no ground truth available so far. The obtained $\bar{F}$-measure is shown in Table~\ref{table:fmow_performance} along with results reported by Christie~\etal~\cite{fmow2017} and results for our best individual classifiers. As previously reported by Christie~\etal~\cite{fmow2017}, using metadata for classification purposes is better than just using the image information. In Table~\ref{table:fmow_performance} we show that on-line data augmentation has also a big impact, increasing the accuracy of their respective versions without augmentation in more than 2\%. Still, our ensemble can improve the performance even more, with a gap of at least 3\% to all individual classifiers.

\begin{table}[!htb]
  \centering
  \normalsize
  \renewcommand{\arraystretch}{1.2}%
  \setlength\tabcolsep{2.5pt}
  \caption{$\bar{F}$-measure for the FMOW testing subset. Evaluated classifiers may use metadata information ($M$) and/or on-line data augmentation ($O$). $\bar{F}$-measure values for classifiers marked with $\dag$ were reported by Christie~\etal~\cite{fmow2017}.}
  \begin{tabular}{|l||c|} \cline{2-2}
   \multicolumn{1}{c||}{}  & $\bar{F}$-measure \\ \cline{1-2}
   Hydra (DenseNet + ResNet)$_{M,O}$ & \textbf{0.781}  \\ \hline   
   DenseNet$_{M,O}$       &  0.749  \\ 
   ResNet$_{M,O}$         &  0.737  \\ \hline 
   LSTM$_{M}^\dag$      &  0.734  \\ 
   DenseNet$_{M}^\dag$  &  0.722  \\ 
   ResNet$_{M}$           &  0.712  \\ \hline 
   LSTM$^\dag$        &  0.688  \\ 
   DenseNet$^\dag$    &  0.679  \\ \hline
  \end{tabular}
  \label{table:fmow_performance}  
\end{table}

With this performance, we managed to finish the competition among the top ten competitors in the provisional scoreboard (5th place). We then submitted a dockerized version of Hydra, which was evaluated by the FMOW organizers using a sequestered dataset. As we show in Table~\ref{table:fmow_final}, the Hydra ensemble was very stable, with nearly the same performance in the testing subset and sequestered dataset. Thanks to this, our final ranking was third place, with accuracy 1\% lower than the best result achieved by another competitor.

\begin{table}[!htb]
\centering
  \normalsize
  \renewcommand{\arraystretch}{1.1}%
  \setlength\tabcolsep{2.5pt}
  \caption{Final ranking: $\bar{F}$-measure for the top five participants of the FMOW challenge using the available testing subset (Open) and the sequestered dataset (Sequestered).}
  \begin{tabular}{cccc}
\bf Rank & \bf Handle & \bf Open & \bf Sequestered \\ \hline
1 & pfr                & 0.7934 & 0.7917 \\   
2 & jianminsun         & 0.7940 & 0.7886 \\ 
3 & usf\_bulls (Hydra) & 0.7811 & 0.7817 \\ 
4 & zaq1xsw2tktk       & 0.7814 & 0.7739 \\ 
5 & prittm             & 0.7681 & 0.7670 \\
  \end{tabular}
  \label{table:fmow_final}  
\end{table}


  


\begin{figure}[!ht]
   \centering
   \includegraphics[width=0.5\textwidth]{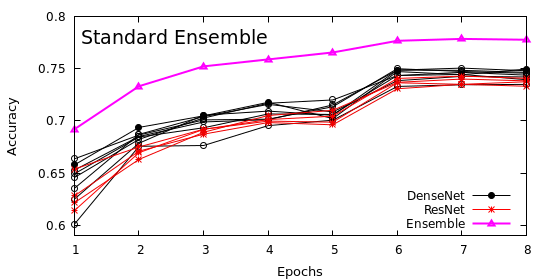}
   \includegraphics[width=0.5\textwidth]{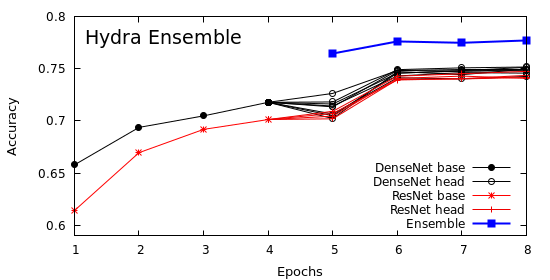}
   \includegraphics[width=0.5\textwidth]{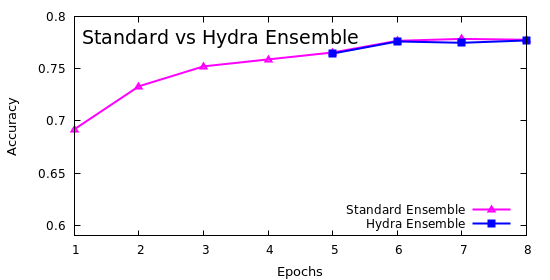}
   \caption{Evaluation of the ensemble accuracy. (top) Standard ensemble, with all classifiers trained from the starting point. Fusion results are shown for each epoch. (middle) The first four epochs show the accuracy of the Hydra's body using DenseNet and Resnet. The following epochs show the accuracy for each head separately as well as for the fusion of all heads. (bottom) Comparison between standard ensemble and Hydra.}
   \label{fig:ensemble_learning}
\end{figure}

To visualize the benefits of using the Hydra framework, we compare it to a standard ensemble that trains all its classifiers from the starting point (ImageNet weights) using the same configuration described in Table~\ref{tab-heads}. The ensemble accuracy over the epochs for this experiment is shown in Figure~\ref{fig:ensemble_learning} together with the accuracy for each classifier. As can be observed, the body of the Hydra provides a good starting point for its heads, which have slightly different accuracy at the end of the training process, very similar to what happens in a standard ensemble. When the heads are combined, the obtained accuracy is considerably higher than the accuracy of the best performing head. The same effect is observed when all classifiers are fused in the standard ensemble, and there is no significant difference in terms of accuracy between Hydra and a standard ensemble.

In terms of computational resources consumption, Hydra has some disadvantages when compared to individual classifiers. If $N$ heads are used, its training and inference costs tend to be $N/2$ and $N$ times more expensive, respectively. However, its accuracy is considerably higher. On the other hand, when compared to the standard ensemble technique with the same number of classifiers and similar accuracy, Hydra's training is approximately two times faster because half of the epochs are carried out for the Hydra's body only.

\revision{Overall, the Hydra framework was a good fit for the FMOW restrictions described in Section~\ref{sec:fmow_restriction}. The time limit for testing was large enough to allow using many classifiers, making the time to train them the main concern. Without Hydra, we would only be able to train half the amount of classifiers, which, according to our experiments, would decrease the results in 1\%. Thus, Hydra succeeded in increasing the number of classifiers for inference without losing their complementarity. As a consequence, we finished the competition among the top competitors.}

\subsection{NWPU-RESISC45 dataset}

We also report the Hydra performance in the NWPU-RESISC45 dataset, which was recently compiled by Cheng~\etal~\cite{7891544}. This dataset has 45 classes, with 700 images per class, each one with a resolution of $256\times256$ pixels, totaling 31,500 images. The intersection of classes between FMOW and NWPU-RESISC45 is reasonable, with some land uses being overcategorized in NWPU-RESISC45 ({\it e.g.}, \textbf{basketball court}, \textbf{baseball diamond}, and \textbf{tennis court} are all labeled as \textbf{recreational facility} in the FMOW dataset) and some classes not representing land use, such as objects ({\it e.g.}, \textbf{airplane} and \textbf{ship}) and vegetation ({\it e.g.}, \textbf{desert}, \textbf{forest}, and \textbf{wetland}). This dataset does not provide satellite metadata.


We followed the experimental setup proposed by Cheng~\etal~\cite{7891544} and randomly chose 10\% or 20\% of the samples in each class for training. The remaining samples were used for testing. We repeated this experiment five times for each configuration and reported the average accuracy and standard deviation. As can be shown in Table~\ref{table:nwpu-resisc45_performance}, Cheng~\etal~\cite{7891544} compared six classifiers based on handcrafted features and three different CNN architectures. In their experiments, the best performance was achieved by a VGG16 network~\cite{Simonyan2015} that was trained using ImageNet weights as a starting point. 

The configuration of the Hydra was slightly different in this experiment. We used 4 DenseNets and 4 ResNets, which were trained for 16 epochs (8 for the body and 8 for the heads). Each head for each architecture has a different augmentation: 1) no augmentation; 2) random vertical and horizontal flips; 3) random zooming; and 4) random shifts. The accuracy results for the NWPU-RESISC45 dataset are shown in Table~\ref{table:nwpu-resisc45_performance}. As NWPU-RESISC45 is less challenging than FMOW, the results are higher and the improvements are lower. The on-line data augmentation increased ResNet and DenseNet accuracies in about 1.0\% and 1.6\%, respectively. Individual classifiers for both architectures outperformed all previous results reported by Cheng~\etal~\cite{7891544}, and the Hydra ensemble was able to achieve the highest accuracy. The improvement on accuracy, however, was slightly greater than 1\% when compared to individual CNNs. It is worth noting that we outperformed the best accuracy results reported by Cheng~\etal~\cite{7891544} by more than 4\%.

\begin{table}[!htb]
  \centering
  \normalsize
  \renewcommand{\arraystretch}{1.1}%
  \setlength\tabcolsep{1.6pt}
  \caption{Accuracy for the NWPU-RESISC45 dataset using training/testing splits of 10\%/90\% and 20\%/80\%. Average accuracy and standard deviation were reported by using five different random splits. Evaluated classifiers may use on-line data augmentation ($O$) and/or ImageNet weights ($I$). Classifiers marked with $\dag$ were reported by Cheng~\etal~\cite{7891544}.} 
  \begin{tabular}{|l||c|c|}
  \cline{2-3}
   \multicolumn{1}{c||}{}  & \multicolumn{2}{c|}{Accuracy} \\  \cline{2-3}
   \multicolumn{1}{c||}{}  & 10\%/90\% & 20\%/80\% \\ \hline
   Hydra (DenseNet + ResNet)$_{O,I}$ & \textbf{92.44} $\pm$ 0.34 & \textbf{94.51} $\pm$ 0.21 \\ \hline
   DenseNet$_{O,I}$        & 91.06 $\pm$ 0.61 & 93.33 $\pm$ 0.55 \\ 
   ResNet$_{O,I}$          & 89.24 $\pm$ 0.75 & 91.96 $\pm$ 0.71 \\ \hline
   VGG16$_{I}$$^\dag$      & 87.15 $\pm$ 0.45 & 90.36 $\pm$ 0.18 \\
   AlexNet$_{I}$$^\dag$    & 81.22 $\pm$ 0.19 & 85.16 $\pm$ 0.18 \\  
   GoogleNet$_{I}$$^\dag$  & 82.57 $\pm$ 0.12 & 86.02 $\pm$ 0.18 \\ \hline
   AlexNet$^\dag$          & 76.69 $\pm$ 0.21 & 79.85 $\pm$ 0.13 \\ 
   VGG16$^\dag$            & 76.47 $\pm$ 0.18 & 79.79 $\pm$ 0.15\\ 
   GoogleNet$^\dag$        & 76.19 $\pm$ 0.38 & 78.48 $\pm$ 0.26\\ \hline
   BoVW$^\dag$             & 41.72 $\pm$ 0.21 & 44.97 $\pm$ 0.28 \\ 
   LLC$^\dag$              & 38.81 $\pm$ 0.23 & 40.03 $\pm$ 0.34 \\ 
   BoVW + SPM$^\dag$       & 27.83 $\pm$ 0.61 & 32.96 $\pm$ 0.47 \\ \hline
   Color histograms$^\dag$ & 24.84 $\pm$ 0.22 & 27.52 $\pm$ 0.14\\    
   LBP$^\dag$              & 19.20 $\pm$ 0.41 & 21.74 $\pm$ 0.18\\  
   GIST$^\dag$             & 15.90 $\pm$ 0.23 & 17.88 $\pm$ 0.22\\ \hline
  \end{tabular}
  \label{table:nwpu-resisc45_performance}  
\end{table}

\section{Conclusions}\label{sec:conclusions}

Automatic and robust classification of aerial scenes is critical for decision-making by governments and intelligence/military agencies in order to support fast disaster response. This research problem, although extensively addressed by many authors, is far from being solved, as evidenced by the final results of the FMOW challenge.

The main contribution of this work was the development of a framework that automatically creates ensembles of CNNs to perform land use classification in satellite images, which is most indicated when the training time is limited. We called this framework Hydra due to its training procedure: the Hydra's body is a CNN that is coarsely optimized and then fine-tuned multiple times to form the Hydra's heads. The resulting ensemble achieved state-of-the-art performance for the NWPU-RESISC45 and the third best performance in the FMOW challenge. Given that there were more than 50 active participants, including teams and individuals from academia and industry, and most of them probably were experimenting different CNN-based ensemble setups, we believe that Hydra is well aligned with the most recent state-of-the-art developments.

\ifCLASSOPTIONcompsoc
  \section*{Acknowledgments}
\else
  \section*{Acknowledgment}
\fi

Part of the equipments used in this project are supported by a grant (CNS-1513126) from NSF-USA . The Titan Xp used for this research was donated by the NVIDIA Corporation.


\ifCLASSOPTIONcaptionsoff
  \newpage
\fi



%

\bibliographystyle{IEEEtran}
\bibliography{paper}

%
\vspace{-20pt}

\begin{IEEEbiography} [{\includegraphics[width=1in,height=1.25in,clip,keepaspectratio]{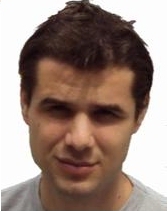}}]{Rodrigo Minetto} is  an  assistant  professor  at  Federal  University  of  Technology - Paran\'{a} (UTFPR) - Brazil. He received the Ph.D. in computer science in 2012 from University of Campinas (UNICAMP), Brazil and Universit\'{e} Pierre et Marie Curie, France (UPMC).  His research interest include image processing, computer vision and machine learning. Currently he is a visiting scholar at University of South Florida (USF), USA.\end{IEEEbiography}

\begin{IEEEbiography}[{\includegraphics[width=1in,height=1.25in,clip,keepaspectratio]{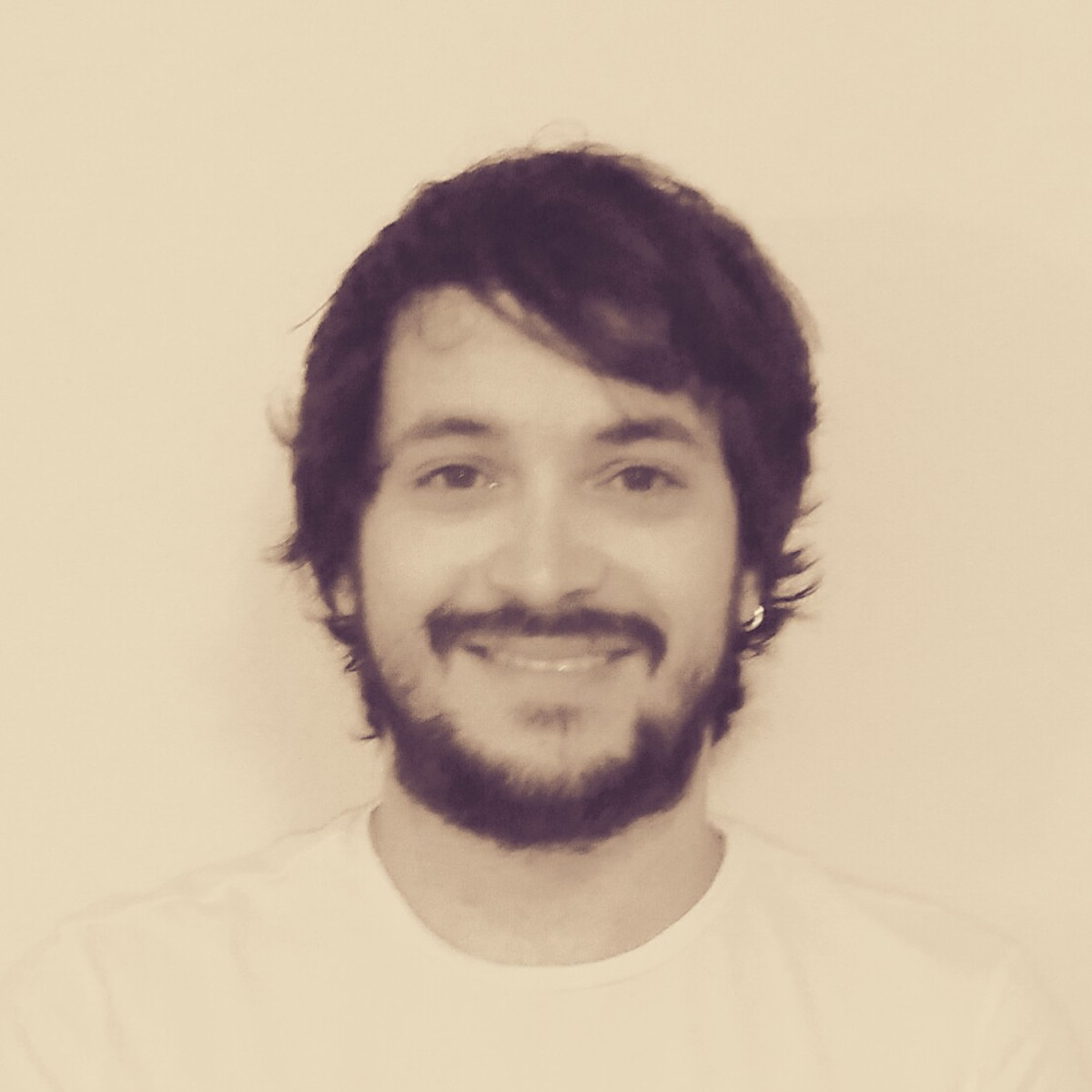}}]{Maur\'icio Pamplona Segundo} is a professor of the Department of Computer Science at the Federal University of Bahia (UFBA). He received his BSc, MSc and DSc in Computer Science from the Federal University of Paran\'{a} (UFPR). He is a researcher of the Intelligent Vision Research Lab at UFBA and his areas of expertise are computer vision and pattern recognition. His research interests include biometrics, 3D reconstruction, accessibility tools, medical images and vision-based automation.\end{IEEEbiography}

\begin{IEEEbiography}[{\includegraphics[width=1in,height=1.25in,clip,keepaspectratio]{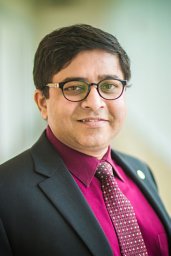}}]{Sudeep Sarkar} is a professor of Computer Science and Engineering and Associate Vice President for Research \& Innovation at the University of South Florida in Tampa. He received his MS and PhD degrees in Electrical Engineering, on a University Presidential Fellowship, from The Ohio State University. He is the recipient of the National Science Foundation CAREER award in 1994, the USF Teaching Incentive Program Award for Undergraduate Teaching Excellence in 1997, the Outstanding Undergraduate Teaching Award in 1998, and the Theodore and Venette Askounes-Ashford Distinguished Scholar Award in 2004. He is a Fellow of the American Association for the Advancement of Science (AAAS), Institute of Electrical and Electronics Engineers (IEEE), American Institute for Medical and Biological Engineering (AIMBE), and International Association for Pattern Recognition (IAPR); and a charter member and member of the Board of Directors of the National Academy of Inventors (NAI). He has 25 year expertise in computer vision and pattern recognition algorithms and systems, holds three U.S. patents and has published high-impact journal and conference papers.\end{IEEEbiography}





\end{document}